\documentclass[conference]{IEEEtran}
\usepackage[bookmarks=false]{hyperref}       
\usepackage{url}            
\usepackage{booktabs}       
\usepackage{amsfonts}       
\usepackage{nicefrac}       
\usepackage{microtype}      
\usepackage{rotating}
\usepackage{hhline}
\usepackage{multirow}
\usepackage{graphicx}
\usepackage{caption}
\usepackage{subcaption}
\usepackage{amsmath,amssymb}
\usepackage{color}

\usepackage{cite}

\usepackage{nicefrac}       
\usepackage{microtype}      

\usepackage{algorithm}
\usepackage{algorithmic}

\usepackage{hhline}
\usepackage{graphicx}
\usepackage{pgfplots}

\DeclareMathOperator*{\argmin}{arg\,min}

\usepackage{footnote}

\begin{document}

\newcommand{\specificthanks}[1]{\@fnsymbol{#1}}
\title{Spatio-Temporal Neural Networks for Space-Time Series Forecasting and Relations Discovery}


\author{Ali~Ziat$^{\dagger}$,
         Edouard~Delasalles$^{\dagger}$,
         Ludovic~Denoyer,
         and~Patrick~Gallinari}
 \author{\IEEEauthorblockN{Ali Ziat$^{* \dagger}$, Edouard Delasalles$^{\dagger}$, Ludovic Denoyer, Patrick Gallinari}
 \IEEEauthorblockA{Sorbonne Universités, UPMC Univ Paris 06, CNRS, LIP6 UMR 7606, 4 place Jussieu 75005 Paris.\\
 $^*$Vedecom Institute, Eco-Mobility Department, 77000, Versailles, France \\
 $^{\dagger}$ Authors have equal contributions.\\
 {\tt\small firstname.name@lip6.fr}}
 }


\maketitle

\begin{abstract}
We introduce a dynamical spatio-temporal model formalized as a recurrent neural network for forecasting time series of spatial processes, i.e. series of observations sharing temporal and spatial dependencies. The model learns these dependencies through a structured latent dynamical component, while a decoder predicts the observations from the latent representations. We consider several variants of this model, corresponding to different prior hypothesis about the spatial relations between the series. The model is evaluated and compared to state-of-the-art baselines, on a variety of forecasting problems representative of different application areas: epidemiology, geo-spatial statistics and car-traffic prediction. Besides these evaluations, we also describe experiments showing the ability of this approach to extract relevant spatial relations.
\end{abstract}

\section{Introduction}
Time series exhibiting spatial dependencies are present in many domains including ecology, meteorology, biology, medicine, economics, traffic, and vision. The observations can come from multiple sources e.g. GPS, satellite imagery, video cameras, etc. Two main difficulties when modeling spatio-temporal data come from their size - sensors can cover very large space and temporal lags - and from the complexity of the data generation process. Reducing the dimensionality and uncovering the underlying data generation process naturally leads to consider latent dynamic models. This has been exploited both in statistics \cite{cressie2011} and in machine learning (ML) \cite{bahadori2014fast,koppula2013learning}.

Deep learning has developed a whole range of models for capturing relevant information representations for different tasks and modalities. For dynamic data, recurrent neural networks (RNN) handles complex sequences for tasks like classification, sequence to sequence prediction, sequence generation and many others \cite{Bengio2008, chung2015recurrent, li2015gated}.

These models are able to capture meaningful features of sequential data generation processes, but the spatial structure, essential in many applications, has been seldom considered in Deep Learning. Very recently, convolutional RNNs \cite{NIPS2015_5955,srivastava2015} and video pixel networks \cite{kalchbrenner2017} have been used to handle both spatiality and temporality, but for video applications only.

We explore a general class of deep spatio-temporal models by focusing on the problem of time series forecasting of spatial processes for different types of data. Although the target in this paper is forecasting, the model can easily be extended to cover related tasks like spatial forecasting (kriging \cite{stein2012interpolation}) or data imputation.


The model, denoted Spatio-Temporal Neural Network (STNN), has been designed to capture the dynamics and correlations in multiple series at the spatial and temporal levels. This is a dynamical system model with two components: one for capturing the spatio-temporal dynamics of the process into latent states, and one for decoding these latent sates into actual series observations.

The model is tested and compared to state of the art alternatives, including recent RNN approaches, on spatio-temporal series forecasting problems: disease prediction, traffic forecasting, meteorology and oceanography. Besides a comparative evaluation on forecasting tasks, the ability of the model to discover relevant spatial relations between series is also analyzed.

The paper is organized as follow: in section \ref{rw} we introduce the related work in machine learning and spatio-temporal statistics. The model is presented in section \ref{model} with its different variants. The experiments are described in section \ref{xp} for both forecasting \ref{xp-forecast} and relations discovery \ref{xp-rel}.
 
\section{Related Work}
\label{rw}
The classical topic of time series modeling and forecasting has given rise to an extensive literature, both in statistics and machine learning. In statistics, classical linear models are based on auto-regressive and moving average components. Most assume linear and stationary time dependencies with a noise component \cite{de200625}. In machine learning, non linear extensions of these models based on neural networks were proposed as early as the nineties, opening the way to many other non linear models developed both in statistics and ML, like kernel methods \cite{muller1999using} for instance.

Dynamical state space models, such as recurrent neural networks, have been used for time series forecasting in different contexts since the early nineties \cite{connor1994recurrent}. Recently, these models have witnessed important successes in different areas of sequence modeling problems, leading to breakthrough in domains like speech \cite{graves2013speech}, language generation \cite{Sutskever2011}, translation \cite{cho2014learning} and many others. A model closely related to our work is the dynamic factor graph model \cite{mirowski2009dynamic} designed for multiple series modeling. Like ours, it is a generative model with a latent component that captures the temporal dynamics and a decoder for predicting the series. However, spatial dependencies are not considered in this model, and the learning and inference algorithms are different.

Recently, the development of non parametric generative models has become a very popular research direction in Deep Learning, leading to different families of innovative and promising models. For example, the Stochastic Gradient Variational Bayes algorithm (SGVB) \cite{kingma2013auto} provides a framework for learning stochastic latent variables with deep neural networks, and has recently been used by some authors to model time series \cite{bayer2014learning, Chung2015, krishnan2015deep}. In our context, which requires to model explicitly both spatial and temporal dependencies between multiple time series, variational inference as proposed by such models is still intractable, especially when the number of series grows, as in our experiments.

Spatio-temporal statistics have already a long history \cite{cressie2011,wikle2010general}. The traditional methods rely on a descriptive approach using the first and second-order moments of the process for modeling the spatio-temporal dependencies. More recently, dynamical state space models, where the current state is conditioned on the past have been explored \cite{Wikle2015}. For these models, time and space can be either continuous or discrete. The usual way is to consider discrete time, leading to the modeling of time series of spatial processes. When space is continuous, models are generally expressed by linear integro-difference equations, which is out of the scope of our work. With discrete time and space, models come down to general vectorial autoregressive formulations. These models face a curse of dimensionality in the case of a large number of sources. Different strategies have been adopted to solve this problem, such as embedding or parameter reduction. This leads to model families that are close to the ones used in machine learning for modeling dynamical phenomena, and incorporate a spatial components. An interesting feature of these approaches is the incorporation of prior knowledge inspired from physical models of space-time processes. This  consists in taking inspiration from prior background of physical phenomena, e.g. diffusion laws in physics, and using this knowledge as guidelines for designing dependencies in statistical models. In climatology, models taking into account both temporal and geographical components have also been used such as Gaussian Markov Random Fields\cite{rue2005gaussian} 

In the machine learning domain, spatio-temporal modeling has been seldom considered, even though some spatio-temporal models have been proposed \cite{ceci2017predictive}. \cite{bahadori2014fast} introduce a tensor model for kriging and forecasting. \cite{koppula2013learning} use conditional random fields for detecting activity in video, where time is discretized at the frame level and one of the tasks is the prediction of future activity. Brain Computer Interface (BCI) is another domain for spatio-temporal data analysis with some work focused on learning spatio-temporal filters \cite{Dornhege2005,ren2014convolutional}, but this is a very specific and different topic.

\section{The STNN Model}
\label{model}
\subsection{Notations and Task}
Let us consider a set of $n$ temporal series, $m$ is the dimensionality of each series and $T$ their length \footnote{We assume that all the series have the same dimensionality and length. This is often the case for spatio-temporal problems otherwise this restriction can be easily removed.}. $m=1$ means that we consider $n$ univariate series, while $m>1$ correspond to $n$ multivariate series each with $m$ components. We will denote $X$ the values of all the series between time $1$ and time $T$. $X$ is then a $\mathbb{R}^{ T \times n \times m}$ 3-dimensional tensor, such that $X_{t,i,j}$ is the value of the j-th component of series $i$ at time $t$. $X_t$ will denote a slice of $X$ at time $t$ such that $X_t \in \mathbb{R}^{n \times m}$ denotes the values of all the series at time $t$.

For simplicity, we first present our model in a mono-relational setting. An extension to multi-relational series where different relations between series are observed is described in section \ref{multi-modal-rel}. We consider that the spatial organization of the sources is captured through a matrix $W \in \mathbb{R}^{n \times n}$. Ideally, $W$ would indicate the mutual influence between sources, given as a prior information. In practice, it might be a proximity or similarity matrix between the sources: for geo-spatial problems, this might correspond to the inverse of a physical distance - e.g. geodesic - between sources. For other applications, this might be provided through local connections between sources using a graph structure (e.g. adjacency matrix for connected roads in a traffic prediction application or graph kernel on the web). In a first step, we make the hypothesis that $W$ is provided as a prior on the spatial relations between the series. An extension where weights on these relations are learned is presented in section \ref{learn-spt-rel}. 

We consider in the following the problem of spatial time series forecasting i.e predicting the future of the series, knowing their past. We want to learn a model $f : \mathbb{R}^{ T \times n \times m} \times \mathbb{R}^{n \times n} \rightarrow \mathbb{R}^{\tau \times n \times m}$ able to predict the future at $\tau$ time-steps of the series based on $X$ and on their spatial dependency.

\subsection{Modeling Time Series with Continuous Latent Factors}
\label{mts}
Let us first introduce the model in the simpler case of multiple time series prediction, without considering spatial relations. The model has two components.

The first one captures the dynamic of the process and is expressed in a latent space. Let $Z_t$ be the latent representation, or latent factors, of the series at time $t$. The dynamical component writes $Z_{t+1}=g(Z_t)$. The second component is a decoder which maps latent factors $Z_t$ onto a prediction of the actual series values at $t$: $\tilde{X_t} = d(Z_t)$, $\tilde{X_t}$ being the prediction computed at time $t$. In this model, both the representations $Z_t$ and the parameters of the dynamical and decoder components are learned. Note that this model is different from the classical RNN formulations \cite{hochreiter1997long, cho2014learning}. The state space component of a RNN with self loops on the hidden cells writes $Z_{t+1}=g(Z_t,X'_t)$, where $X'_t$ is the ground truth $X_t$ during training, and the predicted value $\tilde{X_t}$ during inference. In our approach, latent factors $Z_t$ are learned during training and are not an explicit function of past inputs as in RNNs: the dynamics of the series are then captured entirely in the latent space.

This formal definition makes the model more flexible than RNNs since not only the dynamic transition function $g(.)$, but also the state representations $Z_t$ are learned from data. A similar argument is developed in \cite{mirowski2009dynamic}. It is similar in spirit to Hidden Markov models or Kalman filters.
\paragraph{Learning problem}
Our objective is to learn the two mapping functions $d$ and $g$ together with the latent factors $Z_t$, directly from the observed series. We formalize this learning problem with a bi-objective loss function that captures the dynamics of the series in the latent space and the mapping from this latent space to the observations. Let $\mathcal{L}(g,d,Z)$ be this objective function:
\begin{equation}
\begin{gathered}
 \mathcal{L}(d,g,Z) = \frac{1}{T}\sum\limits_t \Delta(d(Z_t),X_t) \hfill \text{ (i)}\\+ \lambda \frac{1}{T} \sum\limits_{t=1}^{T-1} || Z_{t+1} - g(Z_t) ||^2 \hfill \text{ (ii)}
\end{gathered}
\label{eqfold}
\end{equation}

The first term (i) measures the ability of the model to reconstruct the observed values $X_t$ from the latent factor $Z_t$. It is based on loss function $\Delta$ which measures the discrepancy between predictions $d(Z_t)$ and ground truth $X_t$. The second term (ii) aims at capturing the dynamicity of the series in the latent space. This term forces the system to learn latent factors $Z_{t+1}$ that are as close as possible to $g(Z_t)$. Note that in the ideal case, the model converges to a solution where $Z_{t+1}=g(Z_t)$, which is the classical assumption made when using RNNs. The hyper-parameter $\lambda$ is used here to balance this constraint and is fixed by cross-validation. The solution $d^*,g^*,Z^*$ to this problem is computed by minimizing $\mathcal{L}(d,g,Z)$:
\begin{equation}
d^*,g^*,Z^* = \arg \min\limits_{d,g,Z} \mathcal{L}(d,g,Z)
\label{eqf}
\end{equation}

\paragraph{Learning algorithm}
In our setting, functions $d$ and $g$, described in the next section, are differentiable parametric functions. Hence, the learning problem can be solved end-to-end with Stochastic Gradient Descent (SGD) techniques\footnote{In the experiments, we used the Nesterov's Accelerated Gradient (NAG) method \cite{sutskever2013importance}.} directly from \eqref{eqf}. At each iteration, a pair $(Z_t, Z_{t+1})$ is sampled, and $Z_t$, $Z_{t+1}$, $g$ and $d$ are updated according to the gradient of \eqref{eqfold}. Training can also be performed via mini-batch, meaning that for each iteration several pairs $(Z_t, Z_{t+1})$ are sampled, instead of a single pair. This results in a high learning speed-up when using GPUs which are the classical configuration for running such methods.
\paragraph{Inference} Once the model is learned, it can be used to predict future values of the series. The inference method is the following: the latent factors of any future state of the series is computed using the $g$ function, and the corresponding observations is predicted by using $d$ on these factors. Formally, let us denote $\tilde{Z}_\tau$ the predicted latent factors at time $T+\tau$. The forecasting process computes $\tilde{Z}_{\tau}$ by successively applying the $g$ function $\tau$ times on the learned vector $Z_T$:
\begin{equation}
\tilde{Z}_{\tau} = g \circ g \circ ... \circ g(Z_T)
\end{equation}
and then computes the predicted outputs $\tilde{X}_{\tau}$:
\begin{equation}
\tilde{X}_{\tau}=d(\tilde{Z_{\tau}})
\end{equation}

\subsection{Modeling Spatio-Temporal Series}
\label{msts}
Let us now introduce a spatial component in the model. We consider that each series has its own latent representation at each time step. $Z_t$ is thus a $n \times N$ matrix such that $Z_{t,i} \in \mathbb{R}^N$ is the latent factor of series $i$ at time $t$, $N$ being the dimension of the latent space. This is different from approaches like \cite{mirowski2009dynamic} or RNNs for multiple series prediction, where $Z_t$ would be a single vector common to all the series. The decoding and dynamic functions $d$ and $g$ are respectively mapping $\mathbb{R}^{n \times N}$ to $\mathbb{R}^{n \times m}$ and $\mathbb{R}^{n \times N}$ to $\mathbb{R}^{n \times N}$.

The spatial information is integrated in the dynamic component of our model through a matrix $W \in \mathbb{R}_+^{n \times n}$. In a first step, we consider that $W$ is provided as prior information on the series' mutual influences. In \ref{ref}, we remove this restriction, and show how it is possible to learn the weights of the relations, and even the spatial relations themselves, directly from the observed data. The latent representation of any series at time $t+1$ depends on its own latent representation at time $t$ (intra-dependency) and on the representations of the other series at $t$ (inter-dependency). Intra-dependency will be captured through a linear mapping denoted $\Theta^{(0)} \in \mathbb{R}^{N \times N}$ and inter-dependency will be captured by averaging the latent vector representations of the neighboring series using matrix $W$, and computing a linear combination denoted $\Theta^{(1)} \in \mathbb{R}^{N \times N}$ of this average. Formally, the dynamic model $g(Z_t)$ is designed as follow:
\begin{equation}
	Z_{t+1} = h(Z_t \Theta^{(0)}+ W Z_t \Theta^{(1)})
    \label{dynamic}
\end{equation}
Here, $h$ is a non-linear function. In the experiments we set $h=tanh$ but $h$ could also be a more complex parametrized function like a multi-layer perceptron (MLPs)  for example -- see section \ref{xp}. The resulting optimization problem over $d$, $Z$, $\Theta^{(0)}$ and $\Theta^{(1)}$ writes:

\begin{equation}
\begin{gathered}
d^*, Z^*, \Theta^{(0)*}, \Theta^{(1)*} =\\ \argmin_{d,Z,\Theta^{(0)}, \Theta^{(1)}} \frac{1}{T}\sum\limits_t \Delta(d(Z_t),X_t) +\\
\lambda \frac{1}{T} \sum\limits_{t=1}^{T-1} || Z_{t+1} - h(Z_t \Theta^{(0)}+ W Z_t \Theta^{(1)}) ||^2\\
\text{  with } Z_t \in \mathbb{R}^{n \times N}
\end{gathered}
\label{eqm}
\end{equation}

\begin{table*}[ht]
\centering
\begin{tabular}{|c||c||c|c|c|c|c|c|c|} \hline
Dataset & Task & $n$ & $m$ & nb relations & time-step & total length & training length & \#folds  \\ \hline \hline
Google Flu &  Flu trends & 29 & 1 & 1 to 3 & weeks& $\approx$  10 years & 2 years & 50 \\ 
GHO (25 datasets) & Number of deaths & 91 & 1 & 1 to 3 & years& 45 years & 35 years & 5 \\ \hline 
Wind & Wind speed and orientation & 500 & 2 & 1 to 3 & hours& 30 days & 10 days & 20 \\
PST & Temperature & 2520 & 1 & 8  &  months & $\approx$ 33 years& 10 years & 15 \\ 
\hline
Bejing & Traffic Prediction & 5000 & 1 &  1 to 3 & 15 min& 1 week & 2 days & 20 \\ \hline
\end{tabular}

\caption{Datasets statistics. $n$ is the number of series, $m$ is the dimension of each series, $timestep$ corresponds to the duration of one time-step and \textit{\#folds} corresponds to the number of temporal folds used for validation. For each fold, evaluation has been made on the next 5 values at $T+1,T+2,...,T+5$. The relation columns specifies the number of different relation types used in the experiments i.e the number of $W^{(r)}$ matrices used in each dataset. 1 to 3 means that the best among 1 to 3 relations was selected using cross validation}
\label{datasets}
\end{table*}

\subsection{Modeling different types of relations}
\label{multi-modal-rel}
The model in section \ref{msts} considers that all the spatial relations are of the same type (e.g. source proximity). For many problems, we will have to consider different types of relations. For instance, when sensors correspond to physical locations and the target is some meteorological variable, the relative orientation or position of two sources may imply a different type of dependency between the sources. In the experimental section, we consider problems with relations based on the relative position of sources, $north, south, west, east, ...$. The multi-relational framework generalizes the previous formulation of the model, and allows us to incorporate more abstract relations, like different measures of proximity or similarity between sources. For instance, when sources are spatiality organized in a graph, it is possible to define different graph kernels, each one of them modeling a specific similarity. The following multi-relational formulation is based on adjacency matrices, and can directly incorporate different graph kernels. 



\begin{figure}[ht]
  \begin{center}
   	\includegraphics[width=1\linewidth]{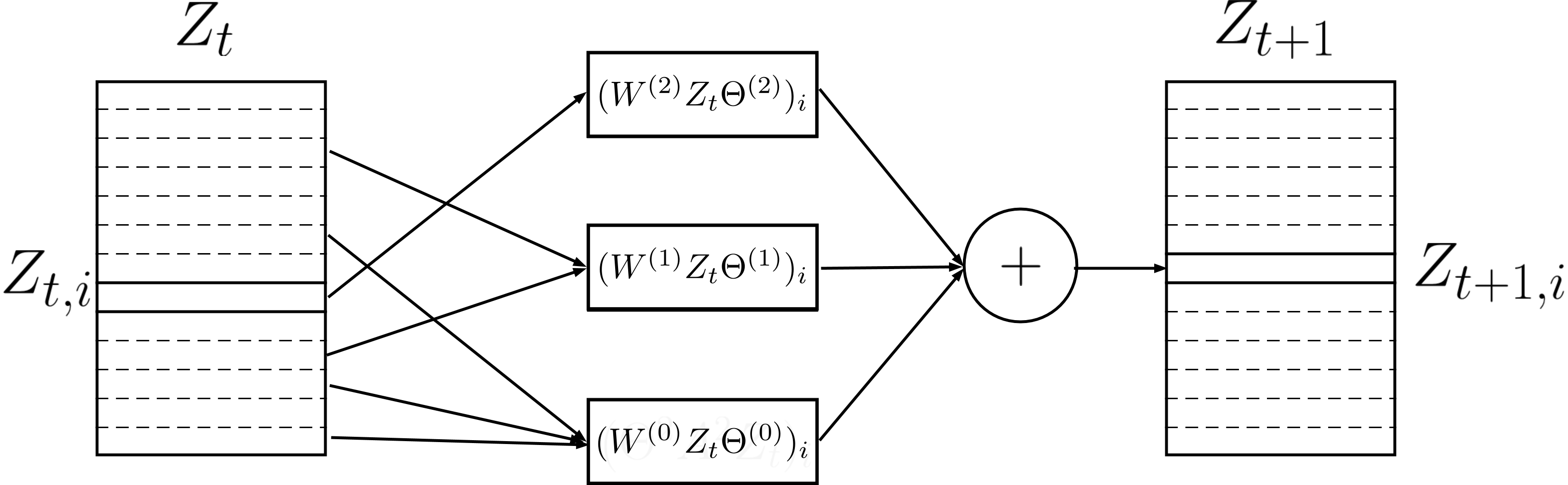} \\
  \end{center}
  \caption{Architecture of the STNN model as described in Section \ref{multi-modal-rel}}.
  \label{networks}
\end{figure}

Each possible relation type is denoted \textit{r} and is associated to a matrix $W^{(r)} \in \mathbb{R}_+^{n \times n}$. For now, and as before, we consider that the $W^{(r)}$ are provided as prior knowledge. Each type of relation \textit{r} is associated to a transition matrix $\Theta^{(r)}$. This learned matrix captures the spatio-temporal relationship between the series for this particular type of relation. The model dynamics writes:
\begin{equation}
Z_{t+1} = h( Z_t \Theta^{(0)}+\sum\limits_{r \in \mathcal{R}} W^{(r)} Z_t \Theta^{(r)})
\label{dynamic-multi-rel}
\end{equation}
where $\mathcal{R}$ is the set of all possible types of relations. The learning problem is similar to equation \eqref{eqm} with $Z_{t+1}$ replaced by the expression in \eqref{dynamic-multi-rel}. The corresponding model is illustrated in figure \ref{networks}. This dynamic model aggregates the latent representations of the series for each type of relation, and then applies $\Theta^{(r)}$ on this aggregate. Each $\Theta^{(r)}$ is able to capture the dynamics specific to relation $(r)$.

\section{Learning the relation weights and capturing spatio-temporal correlations}
\label{learn-spt-rel}
\label{ref}
In the previous sections, we made the hypothesis that the spatial relational structure and the strength of influence between series were provided to the model through the $W^{(r)}$ matrices. We introduce below an extension of the model where weights on these relations are learned. This model is denoted STNN-R(efining). We further show that, with a slight modification, this model can be extended to learn both the relations and their weights directly from the data, without any prior. This extension is denoted  STNN-D(iscovering).



\begin{table*}[t]
\centering
\begin{tabular}{|c||c|c|c|c|c|c|}
\hline
\multirow{2}{*}{Models} & \multicolumn{2}{c|}{Disease}& Car Traffic &\multicolumn{2}{|c}{Geographical}\\
\hhline{~------}
& Google Flu & GHO\footnote{Results shown here correspond to the average for all diseases in the dataset. The detail is in the supplementary material for space convenience.} & Beijing &Speed&Direction&PST \\
\hhline{=:=:=:=:=:=:=}
MEAN & .175& .335&.201&0.191&0.225&.258\\\hline
AR &  .101$\pm .004$&$.299 \pm .008$&$.075 \pm .003$&$.082 \pm .005$&0.098$\pm .016$&$.15\pm .002$\\\hline
VAR-MLP &  $.095 \pm .004$& $.291 \pm .004$ & $.07 \pm .002$&$.071 \pm .005$&0.111$\pm 0.14$&$.132\pm .003$\\\hline
DFG &  $.095 \pm .008$& $.288 \pm .002$ & $.068 \pm .005$&$.07 \pm .004$&$.092 \pm .006$&$.99\pm .019$\\\hline
RNN-tanh &  $.082 \pm .008$& $.287 \pm .011$ &   $.075 \pm .006$&$.064 \pm .003$&$.09 \pm .005$&$.141 \pm .01$\\\hline
RNN-GRU &  $.074 \pm .007$ &$.268 \pm .07$&$.074\pm .002$&$.059 \pm .009$&$.083 \pm .005$&$.104\pm .008$\\\hline
STNN &  $.066 \pm .006$ &$\textbf{.261}\pm .009$&$.056\pm .003$&$\textbf{.047}\pm .008$&$\textbf{.061}\pm .008$&$.095\pm .008$\\\hline
STNN-R &  $\textbf{.061} \pm .008$ &$\textbf{.261}\pm .01$&$\textbf{.055}\pm .004$&$\textbf{.047}\pm .008$&$\textbf{.061}\pm .008$&$\textbf{.08}\pm .014$\\\hline
STNN-D &  $.073 \pm .007$ & $.288 \pm .09$&$.069\pm .01$&$.059\pm .008$&$.073\pm .008$&$.109\pm .015$  \\\hline

\end{tabular}
\caption{\label{tab:widgets}Average RMSE for the different datasets computed for T+1, T+2,...,T+5. Standard deviation was computed by re-training the models on different seeds.}
\label{table1}
\end{table*}

\begin{table*}[h]
\begin{center}
\begin{tabular}{|p{4.5cm}|c|c|c|c|c|c|}

\hline
Disease / Model  & AR & VAR-MLP& RNN-GRU &Mean&DFG&STNN-R\\
\hline

All causes&0.237&0.228&0.199&0.35&0.291&\textbf{0.197}\\
\hline
Tuberculosis&0.407&0.418&\textbf{0.37}&0.395&0.421&0.377\\
\hline
Congenital syphilis&0.432&0.443&0.417&0.459&0.422&\textbf{0.409}\\
\hline
Diphtheria&0.406&0.396&0.387&0.404&0.419&\textbf{0.385}\\
\hline
Malignant neoplasm of esophagus&0.355&\textbf{0.341}&\textbf{0.341}&0.363&0.372&0.345\\
\hline
Malignant neoplasm of stomach&0.44&0.434&0.431&0.455&0.452&\textbf{0.43}\\
\hline
&0.267&0.254&0.282&0.303&0.301&\textbf{0.253}\\
\hline
Malignant neoplasm of intestine&0.281&0.29&0.278&0.314&0.305&\textbf{0.275}\\
\hline
Malignant neoplasm of rectum&0.501&0.499&\textbf{0.481}&0.504&0.509&0.498\\
\hline
Malignant neoplasm of larynx&0.321&0.313&0.32&0.314&0.329&\textbf{0.310}\\
\hline
Malignant neoplasm of breast&0.375&0.375&0.382&0.394&0.38&\textbf{0.36}\\
\hline
Malignant neoplasm of prostate&0.111&0.113&\textbf{0.109}&0.184&0.138&\textbf{0.109}\\
\hline
Malignant neoplasm of skin&0.253&0.243&0.227&0.264&0.256&\textbf{0.221}\\
\hline
Malignant neoplasm of bones&0.103&0.099&0.097&0.204&0.173&\textbf{0.08}\\
\hline
Malignant neoplasm of all other and unspecified sites &0.145&0.157&\textbf{0.147}&0.164&0.169&0.156\\
\hline
Lymphosarcoma&0.15&0.132&0.13&0.231&0.135&\textbf{0.122}\\
\hline
Benign neoplasms &0.366&0.362&0.332&0.398&0.331&\textbf{0.331}\\
\hline
Avitaminonsis&0.492&0.474&0.449&0.571&0.58&\textbf{0.414}\\
\hline
Allergic disorders&0.208&0.217&0.221&0.342&0.24&\textbf{0.202}\\
\hline
Multiple sclerosis&0.061&0.057&0.061&0.242&0.152&\textbf{0.056}\\
\hline
Rheumatic fever&0.325&0.31&0.287&0.345&0.313&\textbf{0.256}\\
\hline
Diseases of arteries&0.302&0.301&0.269&0.345&0.328&\textbf{0.238}\\
\hline
Influenza&0.141&0.141&0.155&0.23&0.217&\textbf{0.125}\\
\hline
Pneumonia&0.119&0.128&\textbf{0.1}&0.187&0.187&\textbf{0.1}\\
\hline
Pleurisy&0.246&0.246&0.247&0.29&0.272&\textbf{0.245}\\
\hline
Gastro-enteritis&0.386&0.369&\textbf{0.291}&0.394&0.398&0.295\\
\hline
Disease of teeth&0.344&0.312&0.305&0.413&0.361&\textbf{0.302}\\
\hline
\end{tabular}
\end{center}
\caption{RMSE of STNN-R over the 25 datasets in GHO for T+1, T+2,...,T+5}
\label{fl4}
\end{table*}

We will first introduce the STNN-R extension. Let $\Gamma^{(r)} \in \mathbb{R}^{n \times n}$ be a matrix of weights such that $\Gamma^{(r)}_{i,j}$ is the strength of the relation between series $i$ and $j$ in the relation $r$. Let us extend the formulation in Equation \eqref{dynamic-multi-rel} as follows:
\begin{equation}
Z_{t+1} = h( Z_t \Theta^{(0)}+\sum\limits_{r \in \mathcal{R}} (W^{(r)}\odot \Gamma^{(r)}) Z_t \Theta^{(r)})
\label{gamma-multi-rel}
\end{equation}
where $\Gamma^{(r)}$ is a matrix to be learned, $W^{(r)}$ is a prior i.e a set of observed relations, and $\odot$ is the element-wise multiplication between two matrices. 
 The learning problem can be now be written as:
\begin{equation}
\begin{gathered}
d^*, Z^*, \Theta^*, \Gamma^* =\\ \argmin_{d,Z,\Gamma} \frac{1}{T}\sum\limits_t \Delta(d(Z_t),X_t)+ \gamma |\Gamma| \\+ \lambda \frac{1}{T} \sum\limits_{t=1}^{T-1} || Z_{t+1} - h(\sum\limits_{r \in \mathcal(R)} (W^{(r)} \odot \Gamma^{(r)}) Z_t \Theta^{(r)}) ||^2
\end{gathered}
\label{eqf}
\end{equation}

where $|\Gamma^{(r)}|$ is a $l_1$ regularizing term that aims at sparsifying $\Gamma^{(r)}$. We thus add an hyper-parameter $\gamma$ to tune this regularization factor.

If no prior is available, then simply removing the $W^{(r)}s$ from equation \eqref{gamma-multi-rel} leads to the following model:
\begin{equation}
Z_{t+1} = h( Z_t \Theta^{(0)}+\sum\limits_{r \in \mathcal{R}} \Gamma^{(r)} Z_t \Theta^{(r)})
\label{gamma-multi-rel-no-W}
\end{equation}

where $\Gamma^{(r)}$ is no more constrained by the prior $W^{(r)}$ so that it will represent both the relational structure and the relation weights. Both models are learned with SGD, in the same way as described in \ref{mts}. The only difference is that a gradient step on the $\Gamma^{(r)}s$ is added.

\section{Experiments}
\label{xp}
\begin{figure*}[t]
\begin{center}
\begin{tabular}{ccc}
Ground Truth & RNN-GRU & STNN-R \\
\includegraphics[width=0.3\linewidth]{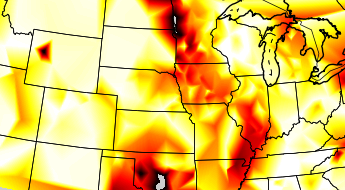} &
\includegraphics[width=0.3\linewidth]{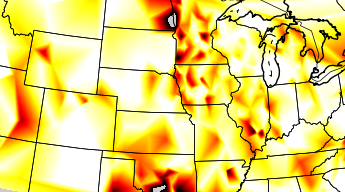} & 
\includegraphics[width=0.3\linewidth]{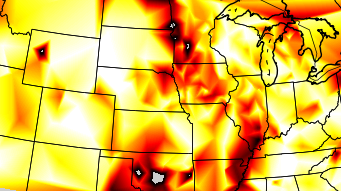} 
\end{tabular}
\end{center}
\caption{Prediction of wind speed over around 500 stations on the US territory. prediction is shown at time-step $T+1$ for RNN-GRU (centre) and STNN-R (right).}
\label{meteo}
\end{figure*}

\begin{figure*}
  \centering
  \begin{tabular}{ccc}
     Ground Truth & RNN-GRU & STNN-R \\
     \includegraphics[width=0.30\linewidth]{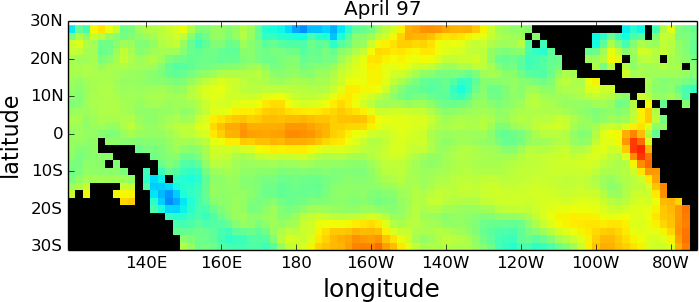} 
      & \includegraphics[width=0.30\linewidth]{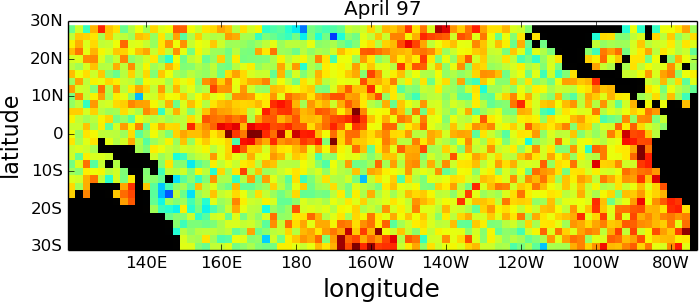}
      & \includegraphics[width=0.30\linewidth]{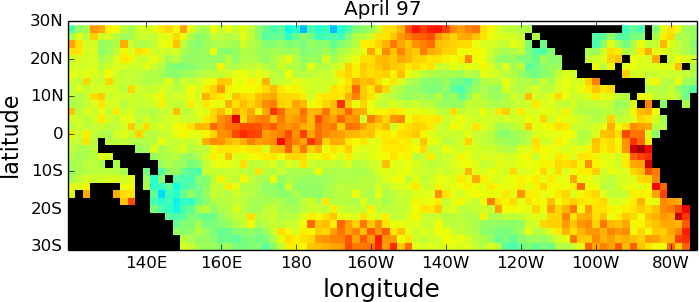}\\ 
   \includegraphics[width=0.30\linewidth]{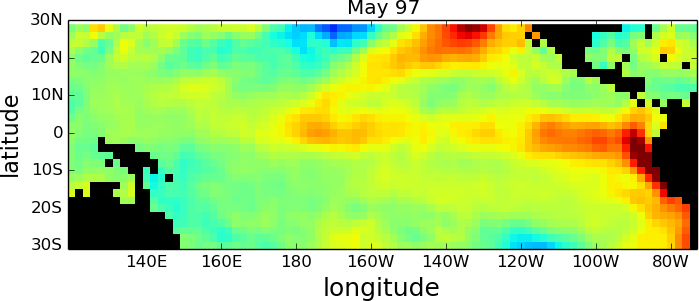} 
      & \includegraphics[width=0.30\linewidth]{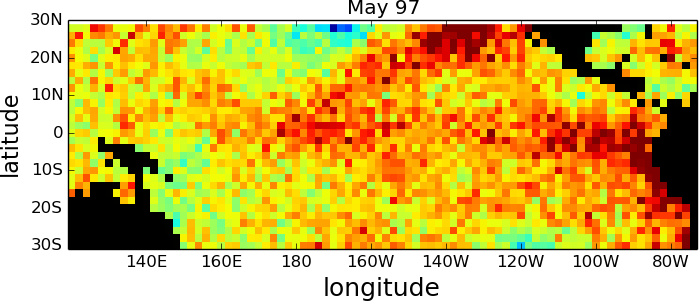}
      & \includegraphics[width=0.30\linewidth]{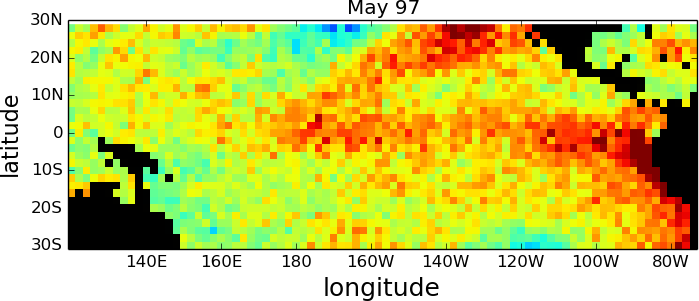}\\ 
         \includegraphics[width=0.30\linewidth]{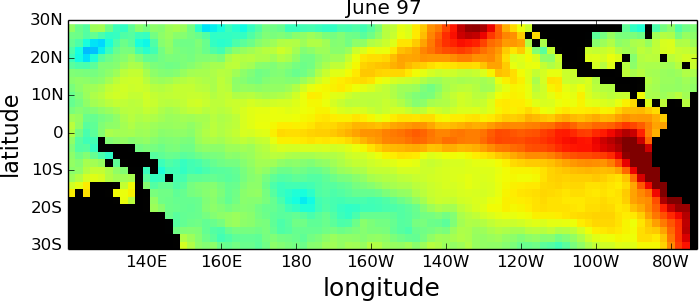} 
      & \includegraphics[width=0.30\linewidth]{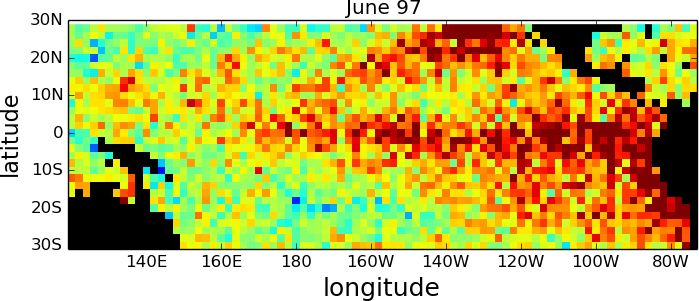}
      & \includegraphics[width=0.30\linewidth]{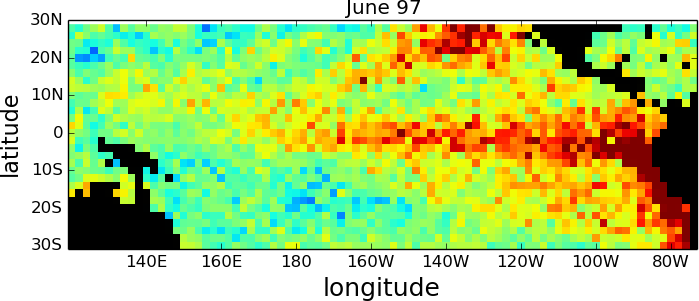}
  \end{tabular}
  \caption{Example of a 3 months prediction of Pacific temperature. Left column is the ground truth, central and right columns correspond respectively to RNN-GRU and STNN-R predictions at horizon $T+1$, $T+2$ and $T+3$ (top to bottom).}
\label{oceano}
\end{figure*}

Experiments are performed on a series of spatio-temporal forecasting problems representative of different domains. We consider predictions within a $+5$ horizon i.e. given a training series of size $T$, the evaluation of the quality of the model will be made over $T+1$ to $T+5$ time steps. The different model hyper-parameters are selected using a time-series cross-validation procedure called rolling origin as in \cite{ben2014boosting,ganeshapillai2013learning}. This protocol makes use of a sliding window of size $T'$: on a series of length $T$, a window of size $T'$  shifted several times in order to create a set of train/test folds. The beginning of the $T'$ window is used for training and the remaining for test. The value of $T'$ is fixed so that it is large enough to capture the main dynamics of the different series. Each series was re-scaled between $0$ and $1$.

We performed experiments with the following models:\\
(i) \textbf{Mean}: a simple heuristic which predicts future values of a series with the mean of its observed past values computed on the $T'$ training steps of each training fold.\\
(ii) \textbf{AR}: a classical univariate Auto-Regressive model. For each series and each variable of the series, the prediction is a linear function of $R$ past lags of the variable, $R$ being a hyper-parameter tuned on a validation set.\\
(iii) \textbf{VAR-MLP}: a vectorial auto-regressive  model where the predicted values of the series at time $t+1$ depend on the past values of all the series for a lag of size $R$. The predictive model is a multi-layer perceptron with one hidden layer. Its performance were uniformly better than a linear VAR. Here again the hidden layer size and the lag $R$ were set by validation\\
(iv) \textbf{RNN-tanh}: a vanilla recurrent neural network with one hidden layer of recurrent units and tanh non-linearities. As for the \textbf{VAR-MLP}, one considers all the series simultaneously, i.e. at time $t$ the RNN receives as input $X_{t-1}$ the values of all the series at $t-1$ and predicts $X_{t}$. A RNN is a dynamical state-space model but its latent state $Z_t$ explicitly depends through a functional dependency both on the preceding values of the series $X_{t-1}$ and on the preceding state $Z_{t-1}$. Note that this model has the potential to capture the spatial dependencies since all the series are considered simultaneously, but does not model them explicitly.\\
(v) \textbf{RNN-GRU}: same as the \textbf{RNN-tanh}, but recurrent units is replaced with gated recurrent units (GRU) units, which are considered state of the art for many sequence prediction problems today \footnote{We also performed tests with LSTM and obtained similar results as GRU.}. We have experimented with several architectures, but using more than one layer of GRU units did not improve the performance, so we used 1 layer in all the experiments.\\
(vi) \textbf{Dynamic Factor Graph (DFG)}: the model proposed in \cite{mirowski2009dynamic} is the closest to ours but uses a joint  vectorial latent representation for all the series as in the RNNs, and does not explicitly model the spatial relations between series.\\
(vii) \textbf{STNN}: our model where $g$ is the function described in equation \eqref{dynamic-multi-rel}, $h$ is the $tanh$ function, and $d$ is a linear function. Note that other architectures for $d$ and $g$ have been tested (e.g. multi-layer perceptrons) without improving the quality of the prediction. The $\lambda$ value has been set by cross validation.\\
(viii and ix) \textbf{STNN-R} and \textbf{STNN-D}: For the forecasting experiments the $\gamma$ value of the $L_1$ penalty  (see equation \eqref{eqf}) wer set to $0$ since higher value decreased the performance, a phenomena often observed in other models such as $L_1$-regularized SVMs. The influence of $\gamma$ on the discovered spatial structure is further discussed and illustrated in figure \ref{sparsity}. 

The complete set of hyper-parameters values for the different models is given in appendix. 

\subsection{Datasets}
\label{datasets}
\begin{figure}[t]
\centering
\begin{tikzpicture}
  \begin{axis}[
      ybar=2pt,
      width=7cm,,
      enlargelimits=0.25,
      legend style={at={(0.5,-0.15)},
        anchor=north,legend columns=-1},
      ylabel={RMSE },
      symbolic x coords={0.0001,0.001,0.01,0.1,1,10,100},
      xtick=data,
      ticklabel style = {font=\tiny},
      nodes near coords align={vertical},
      ]
  \addplot coordinates {(0.0001,0.034) (0.001,0.026) (0.01,0.025) (0.1,0.021) (1,0.027) (10,0.031) (100,0.040)};
  \legend{$\lambda$}
  \end{axis}
\end{tikzpicture}
\caption{RMSE on Google Flu w.r.t $\lambda$}
\label{fl1}
\end{figure}
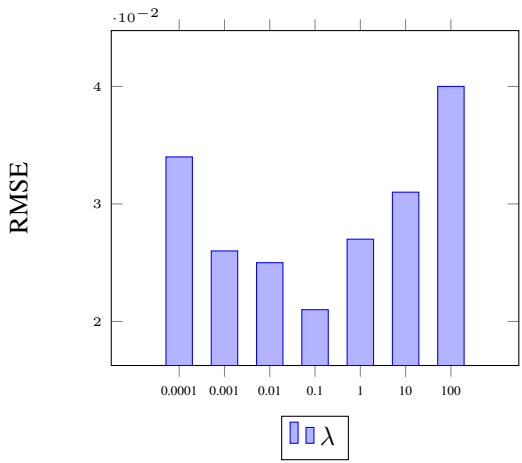

The different forecasting problems and the corresponding datasets are described below. The dataset characteristics are provided in table \ref{datasets}.

\begin{figure}
\begin{center}
\includegraphics[width=0.9\linewidth]{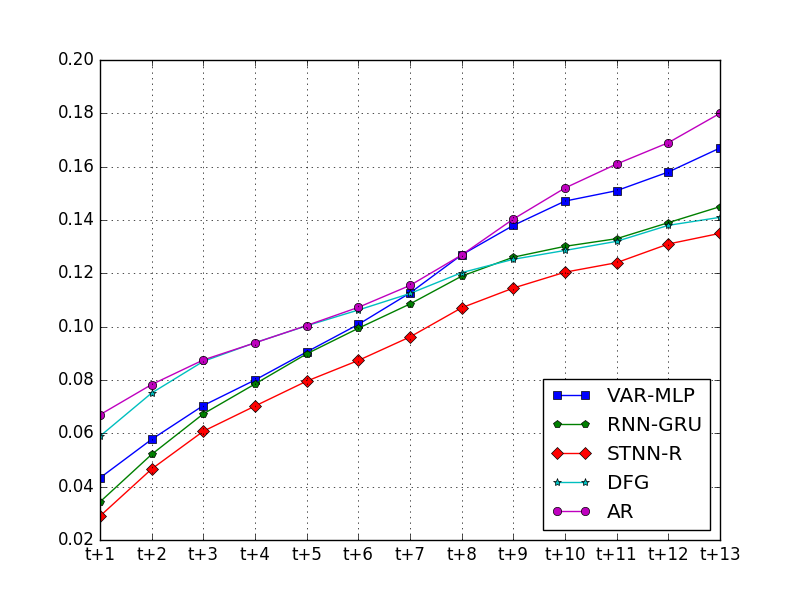}
\end{center}
\caption{RMSE on the Google Flu dataset at horizon $T+1$ to $T+13$}
\label{curve}
\end{figure}

\begin{itemize}
\item \textbf{Disease spread forecasting:} The \textbf{Google Flu} dataset contains for 29 countries, about ten years of weekly estimates of influenza activity computed by aggregating Google search queries (see \url{http://www.google.org/flutrends}). We extract binary relations between the countries, depending on whether or not they share a border, as a prior $W$.
\item \textbf{Global Health Observatory (GHO): } This dataset made available by the Global Health Observatory, (\url{http://www.who.int/en/}) provides the number of deaths for several diseases. We picked 25 diseases corresponding to \textbf{25 different datasets}, each one composed of 91 time series corresponding to 91 countries (see table \ref{datasets}). Results are averages over all the datasets. As for Google Flu, we extract binary relations $W$ based on borders between the countries.
\item \textbf{Geo-Spatial datasets:} The goal is to predict the evolution of geophysical phenomena measured on the surface of the Earth. \\The \textbf{Wind} dataset (\url{www.ncdc.noaa.gov/}) consists of hourly summaries of meteorological data. We predict wind speed and orientation for approximately 500 land stations on U.S. locations. In this dataset, the relations correspond to a thresholded spatial proximity between the series.  Given a selected threshold value $d$, two sources are connected ($w_{i,j}=1$) if their distance is below $d$ and not connected ($w_{i,j}=0$) otherwise. \\The \textbf{Pacific Sea Temperature (PST)} dataset represents gridded (at a 2 by 2 degrees resolution, corresponding to 2520 spatial locations) monthly Sea Surface Temperature (SST) on the Pacific for 399 consecutive months from January 1970 through March 2003. The goal is to predict future temperatures at the different spatial locations. Data were obtained from the Climate Data Library at Columbia University (\url{http://iridl.ldeo.columbia.edu/}). Since the series are organized on a 2D grid, we extract 8 different relations : one for each cardinal direction (north, north-west, west, etc...). For instance, the relation $north$, is associated to a binary adjacency matrix $W^{(north)}$ such that $W^{(north)}_{i,j}$ is set to $1$ if and only if source $j$ is located 2 degree at the north of source $i$ (the pixel just above on the satellite image).
\item \textbf{Car Traffic Forecasting:} The goal is to predict car traffic on a network of streets or roads. We use the \textbf{Beijing dataset } provided in \cite{yuan2011driving,yuan2010t} which consists of GPS trajectories for $\sim 10500$ taxis during a week, for a total of 17 millions of points corresponding to road segments in Beijing. From this dataset, we extracted the traffic-volume aggregated on a 15 min window for 5,000 road segments. The objective is to predict the traffic at each segment. We connect two sources if they correspond to road segments with a shared crossroad.
\end{itemize}

For all the datasets but PST (i.e. Google Flu, GHO, Wind and Bejing), we defined the relational structure using a simple adjacency matrix $W$. Based on this matrix, we defined $K$ different relations by introducing the powers of this matrix: $W^{(1)}=W$, $W^{(2)}=W \times W$, etc. In our setting $K$ took values from 1 to 3 and the optimal value for each dataset has been selected during the validation process. 


\subsection{Forecasting Results}
\label{xp-forecast}

\begin{figure}[t]
\begin{center}
\begin{tabular}{c}
\includegraphics[width=0.75\linewidth]{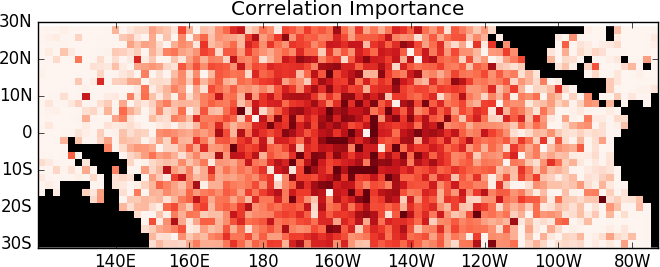} \\
\includegraphics[width=0.75\linewidth]{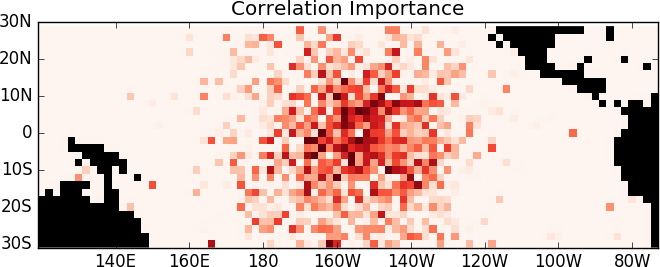} \\
\includegraphics[width=0.75\linewidth]{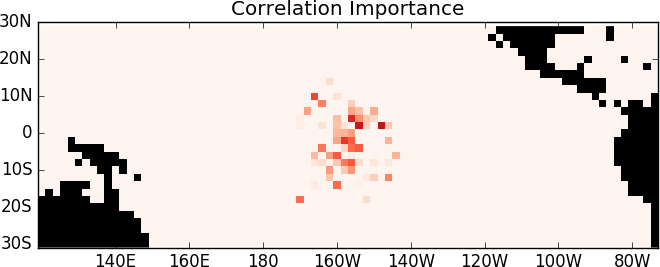} 
\end{tabular}
\end{center}
\caption{Illustrations of correlations $\Gamma$ discovered by the STNN-D model, with $\gamma$ in $\{0.01,0.1,1\}$ ( from top to bottom).}
\label{sparsity}
\end{figure}

\begin{figure}[t]
\includegraphics[width=\linewidth]{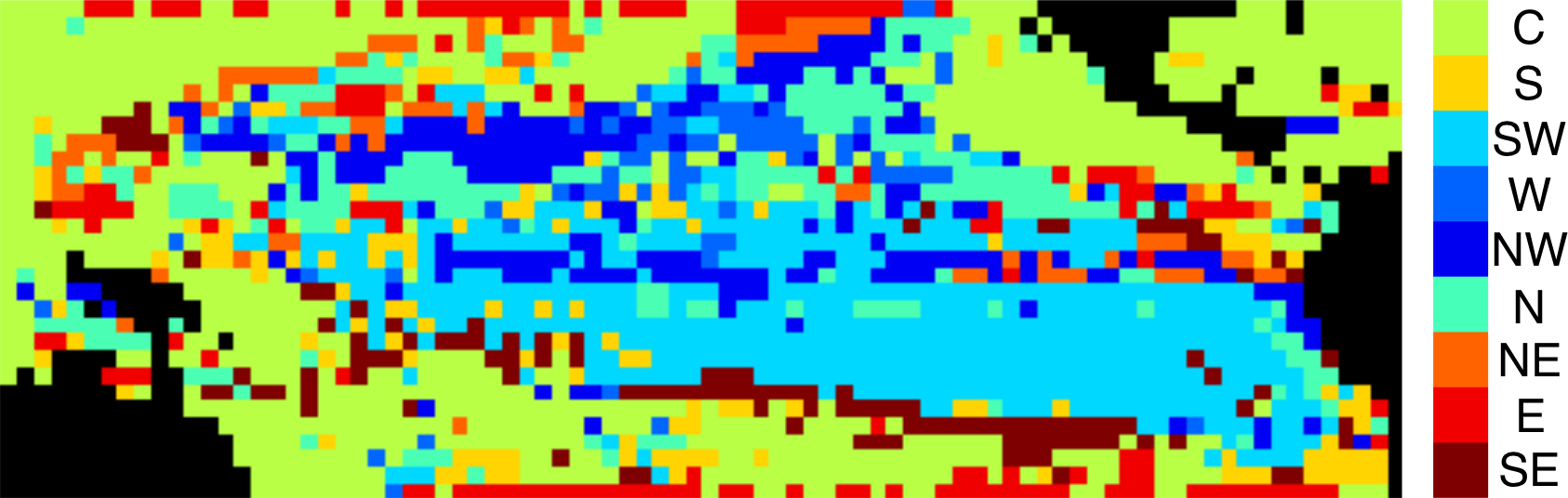}
\caption{Spatial correlations extracted by the STNN-R model on the PST dataset. The color of each pixel correspond to the principal relation extracted by the model.}
\label{alpah_refined_pst}
\end{figure}


A quantitative evaluation of the different models and the baselines, on the different datasets is provided in table \ref{table1}. All the results are average prediction error for $T+1$ to $T+5$ predictions. The score function used is the Root Mean Squared Error (RMSE). A first observation is that STNN and STNN-R models which make use of prior spatial information significantly outperform all the other models on all the datasets. For example, on the challenging PST dataset, our models increase by $23\%$ the performance of the GRU-RNN baseline. The increase is more important when the number of series is high (geo-spatial and traffic datasets) than when it is small (disease datasets). In these experiments, STNN-D is on par with RNN-GRU. The two models do not use prior information on spatial proximity. STNN makes use of a more compact formulation than RNN-GRU for expressing the series mutual dependency but the results are comparable. Vectorial AR logically improves on mono-variable AR (not shown here) and non linear MLP-VAR improves on linear VAR.

\begin{figure}[h]
\centering
\includegraphics[width=0.4\textwidth]{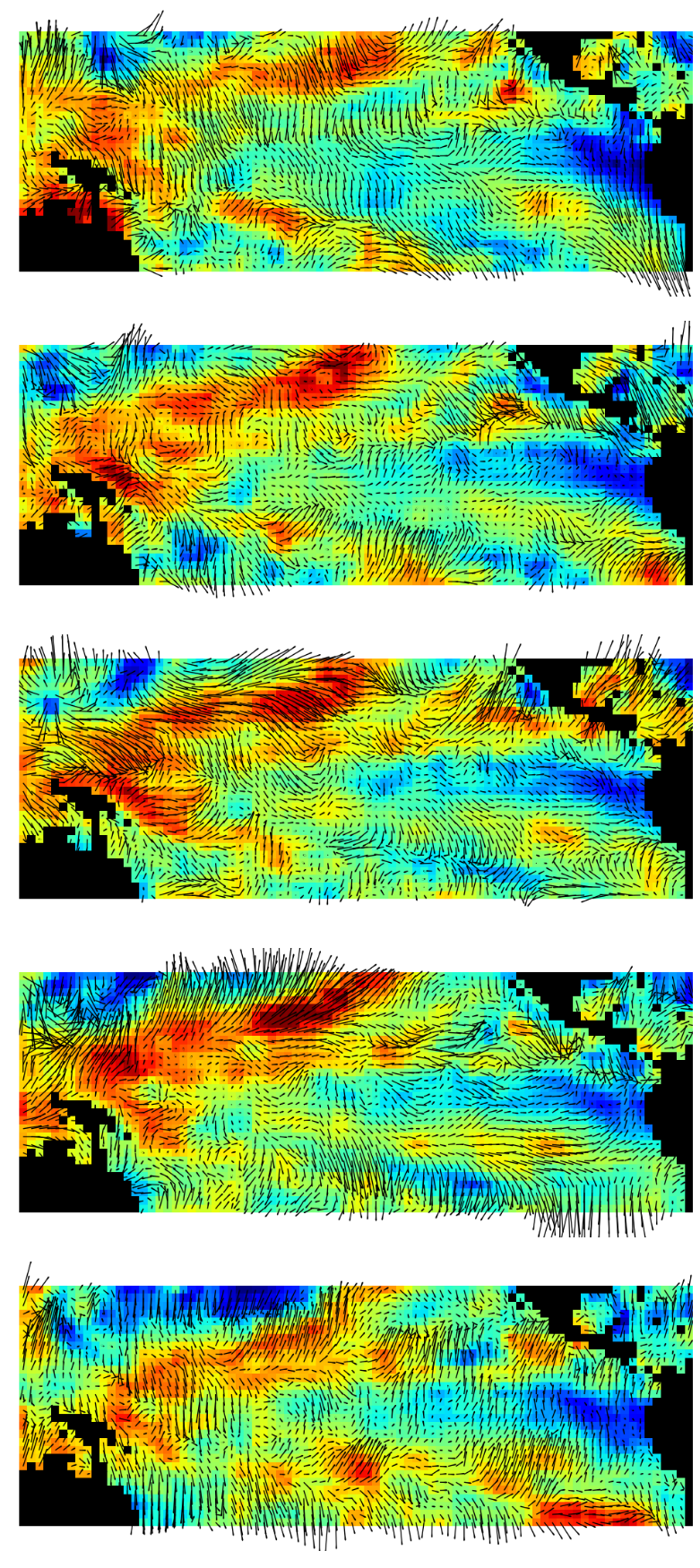}
\caption{Dynamic spatio-temporal relations extract from the PST dataset on the training set. The color represents the actual sea surface temperature. The arrows represent the extracted spatial relations that evolve through time.}
\label{arrows}
\end{figure}

We also provide in table \ref{fl4} the score for each of the 25 diseases of the GHO dataset. STNN-R obtained the best performance compared to STNN and STNN-D. It outperforms state-of-the-art methods in 20 over 25 datasets, and is very close to the RNN-GRU model on the 5 remaining diseases where RNN-GRU performs best. It thus shows that our model is able to benefit from the neighbour information in the proximity graph. 

Figures \ref{meteo} and \ref{oceano} illustrate the prediction of \textit{STNN-R} and \textit{RNN-GRU} on the meteorology and on the oceanography datasets along with the ground truth. Clearly on these datasets, STNN qualitatively performs much better than RNNs by using explicit spatial information. STNN is able to predict fine details corresponding to local interactions when RNNs produce a much more noisy prediction. These illustrations are representative of the general behavior of the two models.

We also provide the performance of the models at different prediction horizons $T+1, T+2,.... T+13$ on figure \ref{curve} for the Google Flu dataset. Results show that STNN performs better than the other approaches for all prediction horizons and is thus able to better capture longer-term dependencies.

Figure \ref{fl1} illustrates the RMSE of the STNN-R model when predicting at $T+1$ on the Google Flu dataset for different values of $\lambda$. One can see that the best performance is obtained for an average value of $\lambda$: low values corresponding to weak temporal constraints do not allow the model to learn the dynamicity of the series while high values degrade the performance of STNN. 

\subsection{Discovering the Spatial Correlations}
\label{xp-rel}

In this subsection, we illustrate the ability of STNN to discover relevant spatial correlations on different datasets. Figure \ref{sparsity} illustrates the values of $\Gamma$ obtained by STNN-D where no structure (e.g. adjacency matrix $W$) is provided to the model on the PST dataset. Each pixel corresponds to a particular time series and the figure shows the correlation $\Gamma_{i,j}$ discovered between each series $j$ with a series $i$ roughly located at the center of the picture. The darker a pixel is, the higher the absolute value of $\Gamma_{i,j}$ is (note that black pixels correspond to countries and not sea). Different levels of sparsity are illustrated from low (up) to high (down). Even if the model does not have any knowledge about the spatial organization of the series (no $W$ matrix provided), it is able to re-discover this spatial organization by detecting strong correlations between close series, and low ones for distant series. 

Figure \ref{alpah_refined_pst} illustrates the correlations discovered on the PST dataset. We used as priors 8 types of relations corresponding to the 8 cardinal directions (South, South-West, etc...). In this case, STNN-R learns weights (i.e $\Gamma^{(r)}$) for each relation based on the prior structure. For each series, we plot the direction with the highest learned weight. The strongest direction for each series is illustrated by a specific color in the figure. For instance, a dark blue pixel indicates that the stronger spatial correlation learned for the corresponding series is the North-West direction. The model extracts automatically relations corresponding to temperature propagation directions in the pacific, providing relevant information about the spatio-temporal dynamics of the system. \\ \linebreak



The model can be adapted to different situations. Figure \ref{arrows} represents the temporal evolution of the spatial relations on the PST dataset. For this experiment, we have  slightly changed the STNN-R model by making the $\Gamma^{(r)}$ time dependent according to:
\begin{equation}
\Gamma^{(r)}_{t,j,i} = f_r(Z_t^i)
\end{equation}
This means that with this modified model, the spatial relation weights depend on the current latent state of the corresponding series and may evolve with time.
In the experiment, $f_r$ is a logistic function. On figure \ref{arrows}, the different plots correspond to successive time steps. The color represent the actual sea surface temperatures, and the arrows represent the direction of the stronger relation weights $\Gamma^{(r)}_{t}$ among the eight possible directions (N, NE, etc). One can see that the model captures coherent dynamic spatial correlations such as global currents directions or rotating motions that gradually evolve with time.




\section{Conclusion}
We proposed a new latent model for addressing multivariate spatio-temporal time series forecasting problems. For this model the dynamics are captured in a latent space and the prediction makes use of a decoder mechanism. Extensive experiments on datasets representative of different domains show that this model is able to capture spatial and temporal dynamics, and performs better than state of the art competing models. This model is amenable to different variants concerning  the formulation of spatial and temporal dependencies between the sources.

For the applications, we have concentrated on forecasting (time based prediction). The same model could be used for interpolating (space based prediction or kriging) or for data imputation when dealing with time-series with missing values.

\section{Acknowledgments}
Locust project ANR-15-CE23-0027-01, funded by Agence Nationale de la Recherche. 
\bibliographystyle{IEEEtran}
\bibliography{IEEEabrv,sample}

\section*{Appendix}
\subsection*{Hyper-parameters selection}
 \label{protocol}
 We list in this section the set of hyper-parameters that have been chosen by cross-validation.
 \begin{itemize}
 \item For \textbf{RNN-models:}
 \begin{itemize}
 \item Size of the hidden state = $(20,50,80,150,300,500)$
 \end{itemize}
 \item For \textbf{AR:}
 \begin{itemize}
 \item Number of lags $R \in \{1,2,5,10,15,25\}$
 \end{itemize}
 \item For \textbf{VAR-MLP:}
 \begin{itemize}
 \item Number of lags $R \in \{1,2,5,10,15,25\}$
 \item Size of the hidden state $ \in \{20,50,80,150,300,500\}$
 \end{itemize}
 \item For \textbf{STNN:}
 \begin{itemize}
 \item Dimension of the latent space $N \in \{5,10,20,50,80,10\}$
 \item Soft-constraint parameter $\lambda \in \{0.001,0.01,0.1,1,10\}$
 \item Sparsity regulation $\gamma \in \{0.001,0.01,0.1,1\}$
 \item $K$ value $\in \{1,2,3\}$ (for datasets composed of 1 unique relation)
 \end{itemize}
 \item For \textbf{DFG:}
 \begin{itemize}
 \item Size of the hidden states $\in \{10,20,50,100,300,500\}$
 \end{itemize}
\end{itemize}

\end{document}